\documentclass[10pt,twocolumn,letterpaper]{article}

\usepackage[final]{cvpr}      

\usepackage{graphicx}
\usepackage{amsmath}
\usepackage{amssymb}
\usepackage{booktabs}
\usepackage{multirow}
\usepackage{xcolor}

\usepackage[pagebackref,breaklinks,colorlinks]{hyperref}

\usepackage[capitalize]{cleveref}
\crefname{section}{Sec.}{Secs.}
\Crefname{section}{Section}{Sections}
\Crefname{table}{Table}{Tables}
\crefname{table}{Tab.}{Tabs.}

\newcommand{\tabref}[1]{Tab.~\ref{#1}}

\newcommand{\figref}[1]{Fig.~\ref{#1}}
\newcommand{\secref}[1]{Sec.~\ref{#1}}

\DeclareMathOperator*{\argmin}{argmin} 

\def\cvprPaperID{8741} 
\def\confName{CVPR}
\def\confYear{2023}

\begin{document}

\title{Offline-to-Online Knowledge Distillation for Video Instance Segmentation}

\author{Hojin Kim, Seunghun Lee, and Sunghoon Im\\
Department of Electrical Engineering \& Computer Science, DGIST, Daegu, Korea\\
{\tt\small \{hojin.kim, lsh5688, sunghoonim\}@dgist.ac.kr}
}

\maketitle

\begin{abstract}

In this paper, we present offline-to-online knowledge distillation (OOKD) for video instance segmentation (VIS), which transfers a wealth of video knowledge from an offline model to an online model for consistent prediction. Unlike previous methods that having adopting either an online or offline model, our single online model takes advantage of both models by distilling offline knowledge. To transfer knowledge correctly, we propose query filtering and association (QFA), which filters irrelevant queries to exact instances. Our KD with QFA increases the robustness of feature matching by encoding object-centric features from a single frame supplemented by long-range global information. We also propose a simple data augmentation scheme for knowledge distillation in the VIS task that fairly transfers the knowledge of all classes into the online model. Extensive experiments show that our method significantly improves the performance in video instance segmentation, especially for challenging datasets including long, dynamic sequences. Our method also achieves state-of-the-art performance on YTVIS-21, YTVIS-22, and OVIS datasets, with mAP scores of 46.1\%, 43.6\%, and 31.1\%, respectively.





\end{abstract}

\section{Introduction}
\label{sec:intro}
\vspace{-2mm}


Video instance segmentation (VIS) is the task of detecting, segmenting, and tracking object instances simultaneously in a given video \cite{yang2019video}. 
It can be categorized into two groups: online and offline approaches. 
Offline methods \cite{wang2021end, cheng2021mask2former, kim2022tubeformer, wu2022seqformer, hwang2021video, lin2021video, bertasius2020classifying, athar2020stem, heo2022vita} input a whole video clip and segment the instances of the entire video sequence in a single step.
These models encode global video knowledge by leveraging detected objects in a video sequence.
This per-clip pipeline generally shows superior performance over per-frame online methods by associating richer information across the entire video sequence.

\begin{figure}[t]
    \centering
    \includegraphics[height=0.55\linewidth]{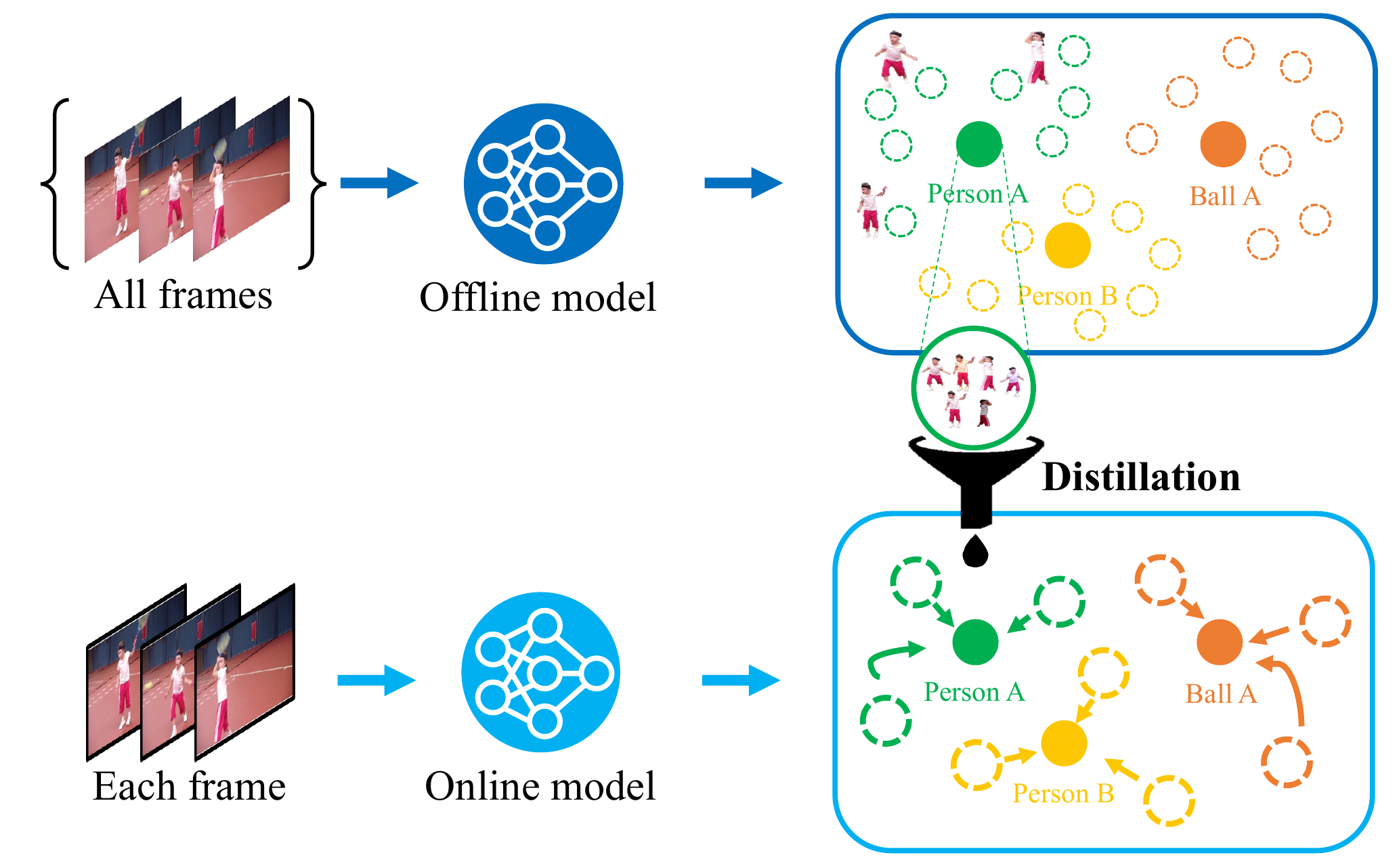}
    \caption{\textbf{Basic concept of our method.} An offline model aggregates the instance-specific features extracted from all frames. 
    We distill the instance feature knowledge encoding global video information into the online model for better instance feature matching.}
    \label{fig:concept}
    \vspace{-2mm}
\end{figure}

Despite the robustness of offline methods, very recent research trends are leaning toward online approaches~\cite{yang2019video,ke2021prototypical, han2022visolo, yang2021crossover, huang2022minvis, cao2020sipmask, fu2021compfeat, li2021spatial}, which segment objects per frame and keep track of instances, using every single video frame as input.
This is useful because many real-time applications (\eg, autonomous driving and surveillance systems) require on-the-fly instance segmentation.
However, they still suffer from inaccurate predictions due to poor matching performance.
This problem occurs because the online method produces inconsistent features for the same instance. 
Moreover, the object-centric feature extraction method makes the matching algorithm to be confused when multiple instances with the same class appear in a frame.
This is the fundamental limitation of online VIS methods and is easily observed in the YouTubeVIS2022 (YTVIS-22) long video benchmark~\cite{xu2022youtubevis}. 

\begin{figure*}
    \centering
    \includegraphics[height=0.35\linewidth]{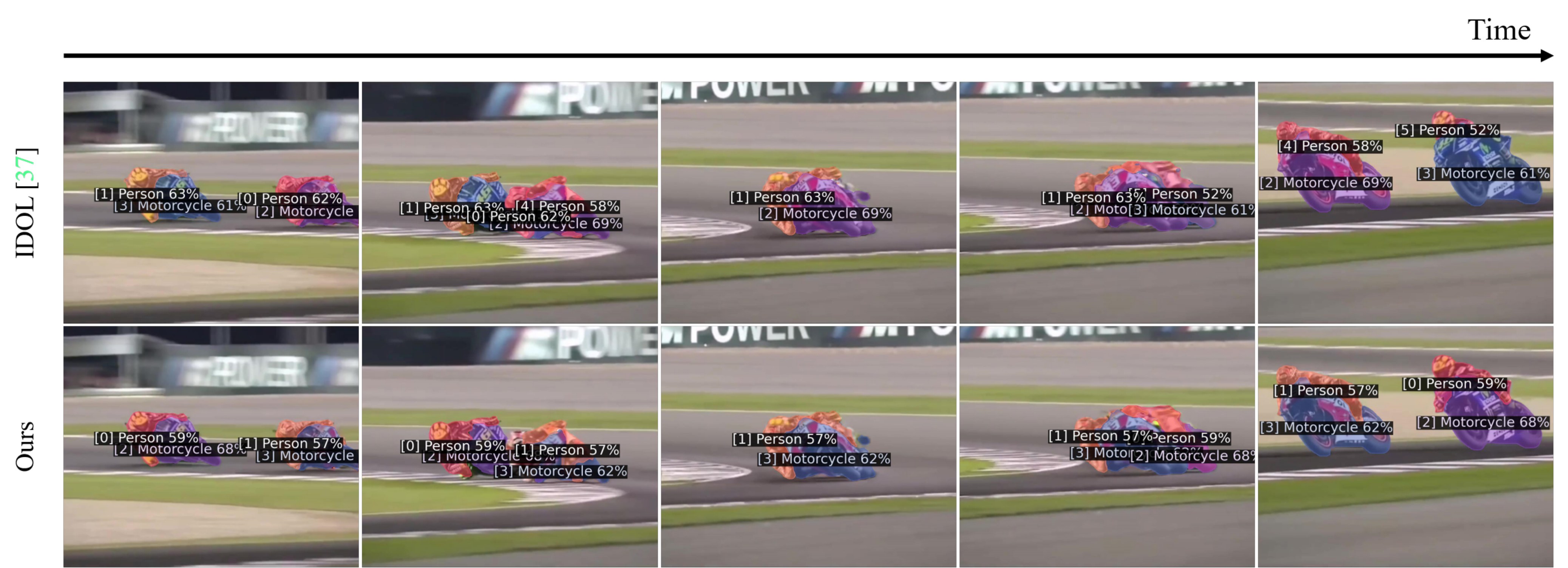}
    \caption{\textbf{Qualitative results for the OVIS dataset.} We compare ours (bottom) with the state-of-the-art online method, IDOL~\cite{wu2022defense} (up).}
    \label{fig:tracking}
\end{figure*}

To tackle these issues while maintaining online applicability, we present Offline-to-Online Knowledge Distillation (OOKD), as shown in \figref{fig:concept}. 
The basic idea behind OOKD is to train an online model
using per-instance features from an offline model as proxy features.
The existing online models~\cite{huang2022minvis,wu2022defense} rely entirely on pair-based loss, which measures the pairwise distance between data in the embedding space.
OOKD encourages our online model to extract instance-specific features embedding global feature knowledge from a single frame.
This allows the online model to learn consistent features, even for dynamic, deformable, and occluded objects, and achieve robust instance matching as shown in \figref{fig:tracking}.

We also propose Query Filtering and Association (QFA) to build high-purity instance features encoded by whole video sequences.
Incorrect predictions of offline models can construct offline knowledge that is less instance discriminative, which leads to performance degradation after the distillation.
The QFA module filters out bad queries and associates instances of the same instance when building offline knowledge from the entire video sequence.
This QFA module is also utilized to link and precisely align the instance features from offline and online models.

Lastly, we find that the VIS task suffers from a class imbalance problem, which makes a model weak at predicting labels from minorities. 
For example, the YTVIS-19 training set contains 1654 `human' class instances out of the total 3774 instances. 
Therefore, we propose a simple yet effective data augmentation for VIS, called Minor-Paste (Minor class copy-Paste).
We adopt the copy-paste scheme~\cite{ghiasi2021simple,kim2022tubeformer}, but selectively sample and paste the instance masks of minor classes.
This module is designed to fairly transfer the knowledge of all classes from the teacher network to the student network.

Extensive experiments show that our method outperforms state-of-the-art methods on all benchmark datasets including YTVIS21~\cite{xu2021youtubevis}, 
YTVIS22~\cite{xu2022youtubevis}, and OVIS~\cite{qi2022occluded}. 
We also observe that our method noticeably improves performance, especially on long video datasets, which validates the effectiveness of the proposed knowledge distillation method and augmentation schemes for VIS.
Our contributions can be summarized as follows:
\begin{itemize}
\setlength\itemsep{0.05em}
    \item To the best of our knowledge, we are the first to introduce an Offline-to-Online Knowledge Distillation (OOKD) method for VIS. 
    \item We design Query Filtering and Association (QFA) for better offline feature extraction and distillation.
    \item We introduce a data augmentation scheme for VIS (Minor-Paste) that allows our student network to learn fairly for all class representations.
    \item The proposed method effectively handles the limitation of online VIS on long videos and outperforms both state-of-the-art online and offline VIS models. 
\end{itemize}

\section{Related Work}
\label{sec:related}

\noindent\textbf{Offline Video Instance Segmentation}
Offline methods~\cite{wang2021end, cheng2021mask2former, kim2022tubeformer, wu2022seqformer, hwang2021video, lin2021video, bertasius2020classifying, athar2020stem} input all scenes at once and predict the instance segmentation labels for all sequence.
VisTR~\cite{wang2021end} is an early work applying the Transformer~\cite{vaswani2017attention} to offline VIS.
IFC~\cite{hwang2021video} presents inter-frame communication Transformers to share inter-frame knowledge with other frames and reduce memory usage.
Some works~\cite{yang2022temporally, hwang2021video} introduce 'near-online' methods that divide the whole video into several clips and use each clip to predict labels.  
VITA~\cite{heo2022vita} uses frame-level object tokens and associates the collection of the features for global video understanding.

\begin{figure*}
  \centering
   \includegraphics[width=1.0\linewidth]{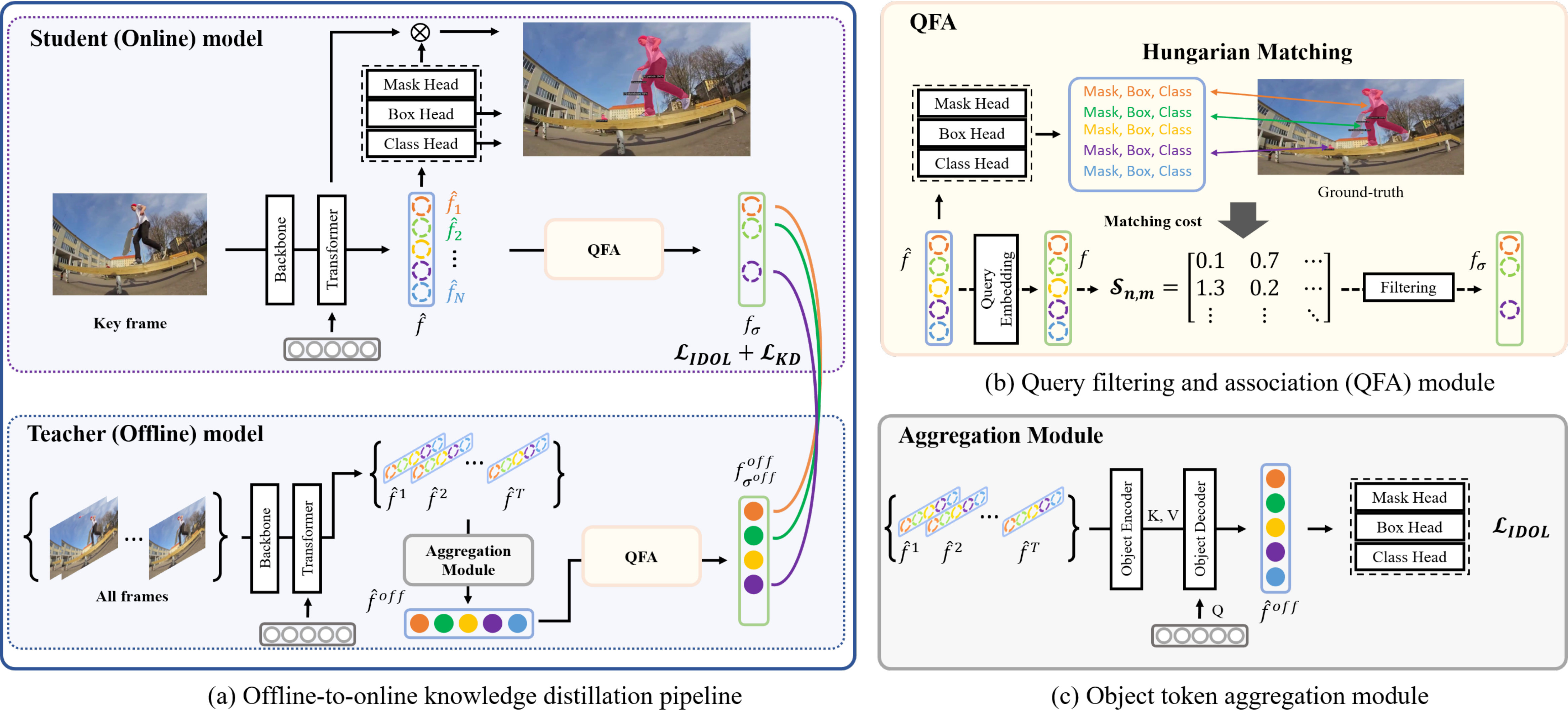}
  \caption{\textbf{Overview of proposed method.} 
  (a) Our pipeline consists of a student and teacher model with the same architecture of backbone network, Transformer (DeformableDETR \cite{zhu2020deformable}) decoder, and query embedding. 
  The student model extracts the features $\{f^t_n\}_{n=\{1,...N\}}$ for each instance $n$ from a single frame $v^t$, which is used as the input of task heads.
  The feature $f^t_n$ also passes the query filtering and association (QFA) module to remove the queries associated with the wrong prediction, as illustrated in (b). 
  The teacher model produces the offline feature $f^{off}_n$ for each instance by aggregating the features $\{\hat{f}^t_n\}^{t=\{1,...,T\}}_{n=\{1,...,N\}}$ for all instances from all the frames, as illustrated in (c).
  Then, we transfer the offline knowledge in the embedded space into online models by imposing a cosine similarity loss $\mathcal{L}_{KD}$ between each instance feature, as well as a task loss $\mathcal{L}_{IDOL}$.
  (b) The QFA module conducts Hungarian Matching \cite{kuhn1955hungarian} between the prediction of each query and the ground-truth label.
  The queries with low matching costs are filtered out.
  (c) The aggregation module unifies the queries of identical objects in the whole video through the object encoder and decoder, which is trained using $\mathcal{L}_{IDOL}$.
  }
  \label{fig:main}
\end{figure*}

\noindent\textbf{Online Video Instance Segmentation} Online methods~\cite{yang2019video,ke2021prototypical, han2022visolo, yang2021crossover, huang2022minvis, cao2020sipmask, fu2021compfeat, li2021spatial} segment instance labels; every single video frame is given as input.
An early online VIS model, MaskTrack R-CNN \cite{yang2019video}, follows the Mask R-CNN \cite{he2017mask} framework with a modified tracking head to match instances between frames.
It becomes a generalized framework, and subsequent works follow the pipeline.
Multi-Object Tracking and Segmentation (MOTS)~\cite{meinhardt2022trackformer, voigtlaender2019mots} is a similar task to online VIS; it predicts segmentation and tracking of all objects except class labels.
MinVIS~\cite{huang2022minvis} trains queries to be discriminative between intra-frame object instances and uses them for instance tracking.
IDOL~\cite{wu2022defense} introduces a memory-based association strategy, which applies contrastive learning to obtain more discriminative instance embeddings.
We adopt the IDOL model as our baseline.



\noindent\textbf{Knowledge Distillation}
Knowledge distillation in neural networks has been widely studied~\cite{hinton2015distilling, gou2021knowledge}.
This technique mainly distills knowledge from a bigger teacher model to a smaller student model.
Recent studies~\cite{ranasinghe2022self, zhao2022progressive, wei2022lidar, hong2022cross} present new concepts of knowledge distillation; this involves the transfer of heterogeneous knowledge learned from a teacher model to a student model. 
SVT~\cite{ranasinghe2022self} introduces a self-distillation method that transfers features from global views to features for local views.
One study on action detection~\cite{zhao2022progressive} proposes to transfer the knowledge from the offline action detection model to an online model.
LiDAR Distillation \cite{wei2022lidar} distills rich knowledge from the LiDAR data with higher beams to lower beams.
The monocular 3D object detection method~\cite{hong2022cross} is trained by distilling feature-based knowledge from 3D LiDAR points.

\noindent\textbf{Data Augmentation}
It is generally known that data augmentation techniques are widely applied in computer vision tasks and are driving performance improvement~\cite{shorten2019survey}.
The augmentation scheme is applied not only to the classifications~\cite{shorten2019survey} but also to the detection~\cite{wang2019data} and segmentation~\cite{ghiasi2021simple}.
Recently, Tubeformer \cite{kim2022tubeformer} extends the copy-paste method~\cite{ghiasi2021simple} to clip-paste for video-level recognition. 
We employ this technique, but more frequently copy and paste the instance mask of the minorities to effectively distill teacher knowledge on all instance representations into the student network.







\section{Method}
\label{sec:method}


First, we briefly describe the overview of IDOL~\cite{wu2022defense}, which is the baseline for our online model in \secref{sec:basemodel}.
Then, we introduce the offline knowledge extraction method in \secref{sec:offknowledge} and the knowledge distillation method in \secref{sec:distillation}.
Lastly, we describe the Minor-Paste data augmentation scheme in \secref{sec:augmentation}.




\subsection{Online VIS model} 
\label{sec:basemodel}


In this paper, we adopt IDOL~\cite{wu2022defense} for a baseline of our online VIS model, which 
consists of an image encoder, a Transformer decoder, and prediction heads. 
Given an input frame $v^t \in \mathbb{R}^{H\times W\times 3}$ of a video $V=\{v^1,...,v^T\}$, either CNN or Transformer backbone extracts feature maps.
The extracted features and $N$ learnable object queries pass through DeformableDETR~\cite{zhu2020deformable} decoder to 
transform the queries into instance features $\{\hat{f}^t_n\}_{n=\{1,...,N\}}$ with $C$ hidden dimension ($\hat{f}^t_n \in \mathbb{R}^{C}$).
Lastly, dynamic mask head~\cite{tian2020conditional} decodes instance features into segmentation mask, bounding box, and a class of instances.
The model is optimized with a classification loss $\mathcal{L}_{cls}$, a bounding box loss $\mathcal{L}_{box}$, a segmentation mask loss $\mathcal{L}_{mask}$, and a contrastive loss $\mathcal{L}_{embed}$ as follows:
\begin{equation}
\begin{aligned}
\mathcal{L}_{\text{IDOL}} &= \mathcal{L}_{cls}+ \lambda_1 \mathcal{L}_{box}+\lambda_2 \mathcal{L}_{mask}+ \lambda_3 \mathcal{L}_{embed}, \\
\end{aligned}
\label{eq:loss}
\end{equation}
where $\lambda_{\{1,2,3\}}$ are the balancing terms among the losses.
Each query $\hat{f}^t_n$ from a frame $v^t$ is the input of the task heads and applied to contrastive embedding to obtain instance embeddings $f^t_n$.
During inference, instance embeddings from previous frames are summed with embeddings in memory banks with specific weights.
Embeddings in memory banks are utilized to match feature similarities between current and memory instances and to track instance IDs.

\subsection{Offline Knowledge Extration}
\label{sec:offknowledge}
We aim to distill instance-distinctive feature knowledge from an offline model into our online model.
We extract the representative features for each instance from the entire video frame and use them as offline knowledge.
For effective distillation, we design the offline model for learning more representative features on the feature space shared with the online model.
Thus, we use the pre-trained online model whose structure is the same as our target online model to extract frame-level instance-centric knowledge. 
We associate the collection of knowledge across an entire video sequence as illustrated in~\figref{fig:main}.

We first pass every single frame $v^t$ into the baseline online VIS model, defined in~\secref{sec:basemodel}, to extract the per-frame instance query $\{\hat{f}^t_n\}_{n=\{1,...,N\}}$.
Then, we aggregate every instance query $\hat{F}_n = \{\hat{f}^t_n\}^{t=\{1,...,T\}}_{n=\{1,...,N\}}$ from the whole set of video frames using object token association \cite{heo2022vita} to obtain offline knowledge $\{\hat{f}^{off}_n\}_{n=\{1,...,N\}}$.
Each instance feature $\hat{f}^{off}_n  \in \mathbb{R}^{C}$ embeds video-level information for each instance.
The aggregation module consists of an object encoder and an object decoder as shown in \figref{fig:main}-(c). 
The encoder builds intercommunication of queries along the temporal axis, employing self-attention modules. 
The augmented instance features and $N$ learnable object queries are passed through the object decoder to embed the offline instance information into the queries. 
The feature aggregation module is trained by passing the offline instance query $\hat{f}^{off}_n$ through the dynamic mask head and minimizing the loss to instance masks, bounding boxes, and classes in \eqref{eq:loss}.
We only train the object encoder and decoder with the frozen baseline online model.
We obtain offline instance embedding $f^{off}_n$ by passing the learned query $\hat{f}^{off}_n$ through contrastive embedding similar to the method in the previous section.

\subsection{Offline-to-Online Knowledge Distillation}
\label{sec:distillation}

Distilling incorrect or irrelevant knowledge into student models can degrade the model's performance.  
This situation is commonly caused by the transfer of knowledge from false predictions in the teacher model and in pairs of mismatched instances between the teacher and student models.
To address the issue, we propose Query Filtering and Association (QFA) in \figref{fig:main}-(b), which maps predicted instance embeddings to ground-truth instances one-to-one while removing wrong predictions.
Suppose we have $M$ ground-truth bounding boxes $B \in \mathbb{R}^{M\times 4}$ and classes $C \in \mathbb{R}^{M \times N_c}$ with $N_c$ class labels in a single frame appearing in a sampled training video.
We define a matching cost matrix $\mathcal{S}\in \mathbb{R}^{N\times M}$ by measuring the localization errors of the bounding box predictions $\hat{B} \in \mathbb{R}^{N \times 4}$ and the confidence errors of the class prediction $\hat{C}\in \mathbb{R}^{N\times N_c}$ as follows:
\begin{equation}
    \mathcal{S}_{n,m} = \mathcal{L}_c (\hat{C}_n,C_m) + \lambda_b \mathcal{L}_b (\hat{B}_n, B_m),
\end{equation}
where $\mathcal{L}_c$ is cross entropy and  $\mathcal{L}_b$ is generalized IoU \cite{rezatofighi2019generalized}, by following \cite{stewart2016end}. The indices $n$ and $m$ are for the prediction and ground truth instances. 
We find the optimal index $\sigma_m$ for ground truth instance $m$, which has the lowest cost among all $N$ predictions, as follows:
\begin{equation}
    \sigma_m = \argmin_{n \in \{1,...,N\}}\mathcal{S}_{n,m}.
\end{equation}






\begin{figure}[t]
    \centering
    \includegraphics[height=0.64\linewidth]{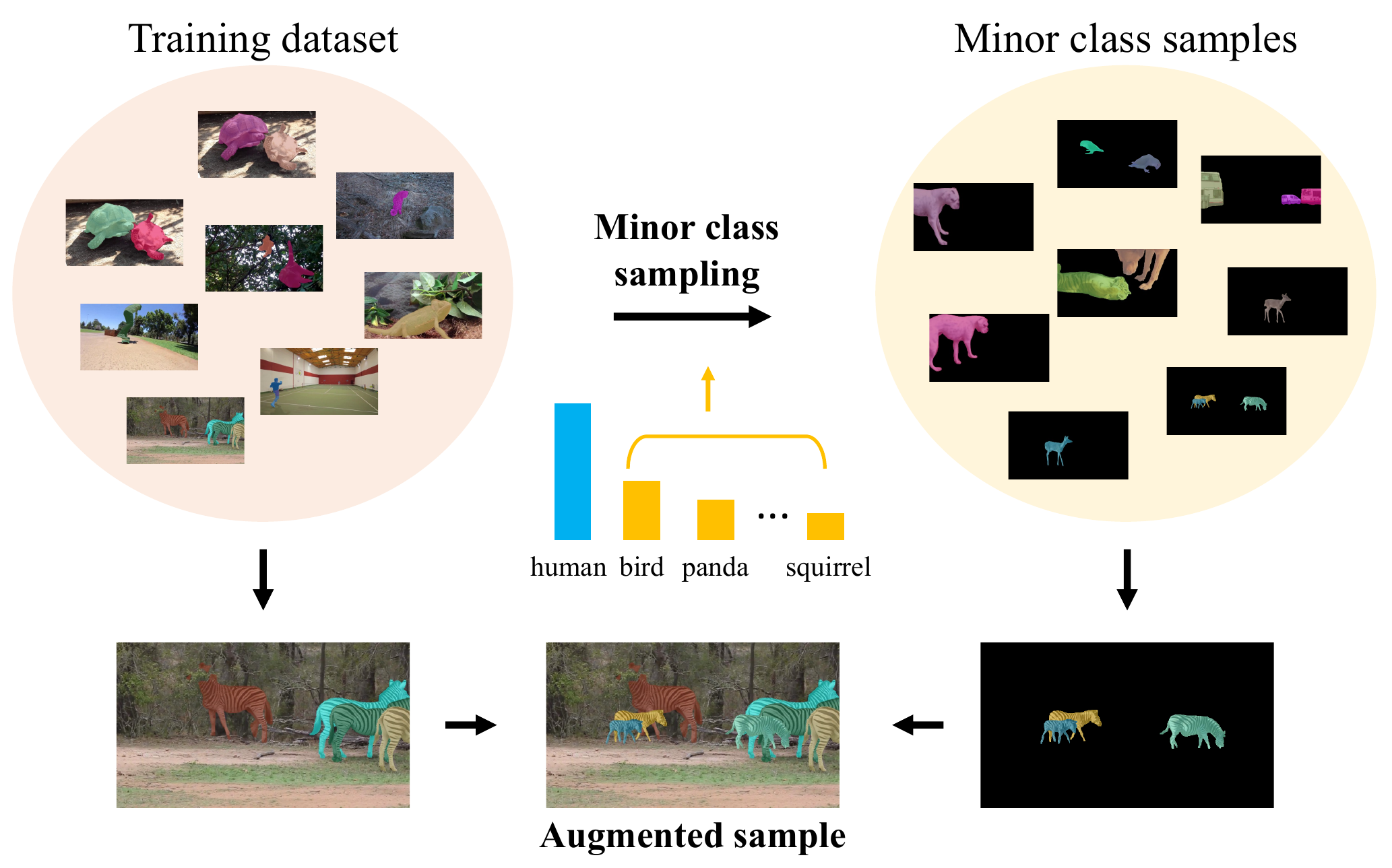}
    \caption{Illustration of our augmentation scheme, Minor-paste.}
    \label{fig:minor}
\end{figure}

We conduct the same process for an offline model to find the optimal index $\sigma^{off}_m$ for offline predictions as well.
These optimal indices $\sigma_m$ and $\sigma^{off}_m$ are utilized to match the instances between online and offline models.
Given pairs of the matched features $f_{\sigma_m}$ and $f^{off}_{\sigma_m^{off}}$, we compute distillation loss, which maximizes the cosine similarity between them, as follows:
\begin{equation}
\mathcal{L}_{KD} = \frac{1}{M} \sum_{m=1}^M  \bigg(1- 
\frac{{f^{off}_{\sigma^{off}_m}}\cdot f_{\sigma_m}}{ \lVert f^{off}_{\sigma^{off}_m}\rVert  \lVert f_{\sigma_m}\rVert}\bigg).
\end{equation}
We additionally impose the distillation loss on the loss with a balancing term $\lambda_4$ defined in \eqref{eq:loss}, as follows:
\begin{equation}
\mathcal{L}_{total} = \mathcal{L}_{\text{IDOL}} + \lambda_{4}\mathcal{L}_{KD}.
\end{equation}


\begin{table*}[h!]
\centering
\begin{tabular}{c | c |  c c c c c c}
 \hline
 Backbone & Type & Method  & mAP & AP$_{50}$ & AP$_{75}$ & AR$_{1}$ & AR$_{10}$ \\ [0.5ex] 
 \hline\hline
 MsgShifT                   &                  Offline & TeViT \cite{yang2022temporally}         & 37.9 & 61.2 & 42.1 & 35.1 & 44.6 \\ \hline
\multirow{15}{*}{ResNet-50} & \multirow{5}{*}{Offline} & VisTR \cite{wang2021end}            & 31.8 & 51.7 & 34.5 & 29.7 & 36.9 \\
                            &                          & IFC \cite{hwang2021video}                 & 36.6 & 57.9 & 39.3 & -    & -    \\
                            &                          & SeqFormer \cite{wu2022seqformer}            & 40.5 & 62.5 & 43.6 & 36.2 & 48.0 \\
                            &                          & Mask2Former \cite{cheng2021mask2former}          & 40.6 & 60.9 & 41.8 & -    & -    \\
                            &                          & VITA \cite{heo2022vita}             & 45.7 & 67.4 & 49.5 & \textbf{40.9} & 53.6 \\ \cline{2-8}
                            & \multirow{10}{*}{Online} & M-RCNN \cite{yang2019video}          & 28.6 & 48.9 & 29.6 & 26.5 & 33.8 \\
                            &                          & STMask \cite{li2021spatial}       & 30.6 & 49.4 & 32.0 & 26.4 & 36.0 \\
                            &                          & SipMask \cite{cao2020sipmask}        & 31.7 & 52.5 & 34.0 & 30.8 & 37.8 \\
                            &                          & Cross-VIS \cite{yang2021crossover}     & 34.2 & 54.4 & 37.9 & 30.4 & 38.2 \\
                            &                          & VISOLO \cite{han2022visolo}                & 36.9 & 54.7 & 40.2 & 30.6 & 40.9 \\
                            &                          & InstanceFormer \cite{koner2022instanceformer}    & 40.8 & 62.4 & 43.7 & 36.1 & 48.1 \\
                            &                          & DeVIS \cite{caelles2022devis}         & 43.1 & 66.8 & 46.6 & 38.0 & 50.1 \\
                            &                          & MinVIS \cite{huang2022minvis}       & 44.2 & 66.0 & 48.1 & 39.2 & 51.7 \\
                            &                          & IDOL \cite{wu2022defense}        & 43.9 & 68.0 & \textbf{49.6} & 38.0 & 50.9 \\
                            &                          & OOKD (Ours)     & \textbf{46.1} & \textbf{69.6} & 49.2 & 40.8 & \textbf{55.5} \\ \hline\hline
\multirow{6}{*}{Swin-L}     & Offline                  & VITA \cite{heo2022vita}       & 57.5 & 80.6 & 61.0 & \textbf{47.7} & 62.6 \\   \cline{2-8}
                            & \multirow{5}{*}{Online}  & InstanceFormer \cite{koner2022instanceformer}  & 51.0 & 73.7 & 56.9 & 42.8 & 56.0 \\
                            &                          & DeVIS \cite{caelles2022devis}        & 54.4 & 77.7 & 59.8 & 43.8 & 57.8 \\
                            &                          & MinVIS \cite{huang2022minvis}        & 55.3 & 76.6 & 62.0 & 45.9 & 60.8 \\
                            &                          & IDOL \cite{wu2022defense}      & 56.1 & 80.8 & 63.5 & 45.0 & 60.1 \\
                            &                          & OOKD (Ours)    & \textbf{59.2} & \textbf{82.6} & \textbf{65.0} & 47.2 & \textbf{64.3} \\
                            
 \hline
\end{tabular}
\caption{Quantative comparison of our method to state-of-the-art methods on the YTVIS-21 dataset. Best scores are highlighted with \textbf{bold}.}
\label{table_YTVIS_2021}
\end{table*}

\subsection{Minor-paste}
\label{sec:augmentation}
We also propose a simple yet effective augmentation scheme, called Minor-paste (minor-class copy-paste) in \figref{fig:minor}, for knowledge distillation in video instance segmentation. 
We aim to transfer the knowledge of the teacher model into the student model equally, regardless of class labels.
To do so, we compute the sampling probability $p^{s}_c$ for each instance $c$ as follows:
\begin{equation}
p^{s}_c = k\frac{\max(p_c)-p_c}{\max(p_c)-\min(p_c)},~\text{where}~c\in \{1,...,N_c\},
\end{equation}
where $p_c$ is the proportion of the number of classes $c$ to the total number of classes in the entire training data set.
We set the scale parameter $k$ at $0.7$, which controls the probability of data augmentation.
We sample video clips containing at least one minor class whose $p_c$ is less than 10\%.
Then, we randomly crop the instance regions and paste them onto other target video clips based on the sampling probability $p^{s}_c$.
We observe that the YTVIS-21 training datasets contains about 35.5\% of the `Human' class and 0.3\% of the `Squirrel' class. Based on these probabilities, for example, the most major class `Human' has never been augmented and the most minor class `Squirrel' has a 70\% chance to be augmented.
We use GT instance segmentation masks and class IDs to determine the area to crop and copy the instance.

%




\section{Experimental Results}
\label{sec:experiments}

\subsection{Experimental Setup}

\noindent\textbf{Datasets:} We examine our method on YTVIS-21 \cite{xu2021youtubevis}, YTVIS-22 \cite{xu2022youtubevis}, and OVIS \cite{qi2022occluded}. These datasets are more recent, and challenging datasets than YTVIS-19 \cite{yang2019video}.

\begin{figure*}
  \centering
   \includegraphics[width=0.89\linewidth]{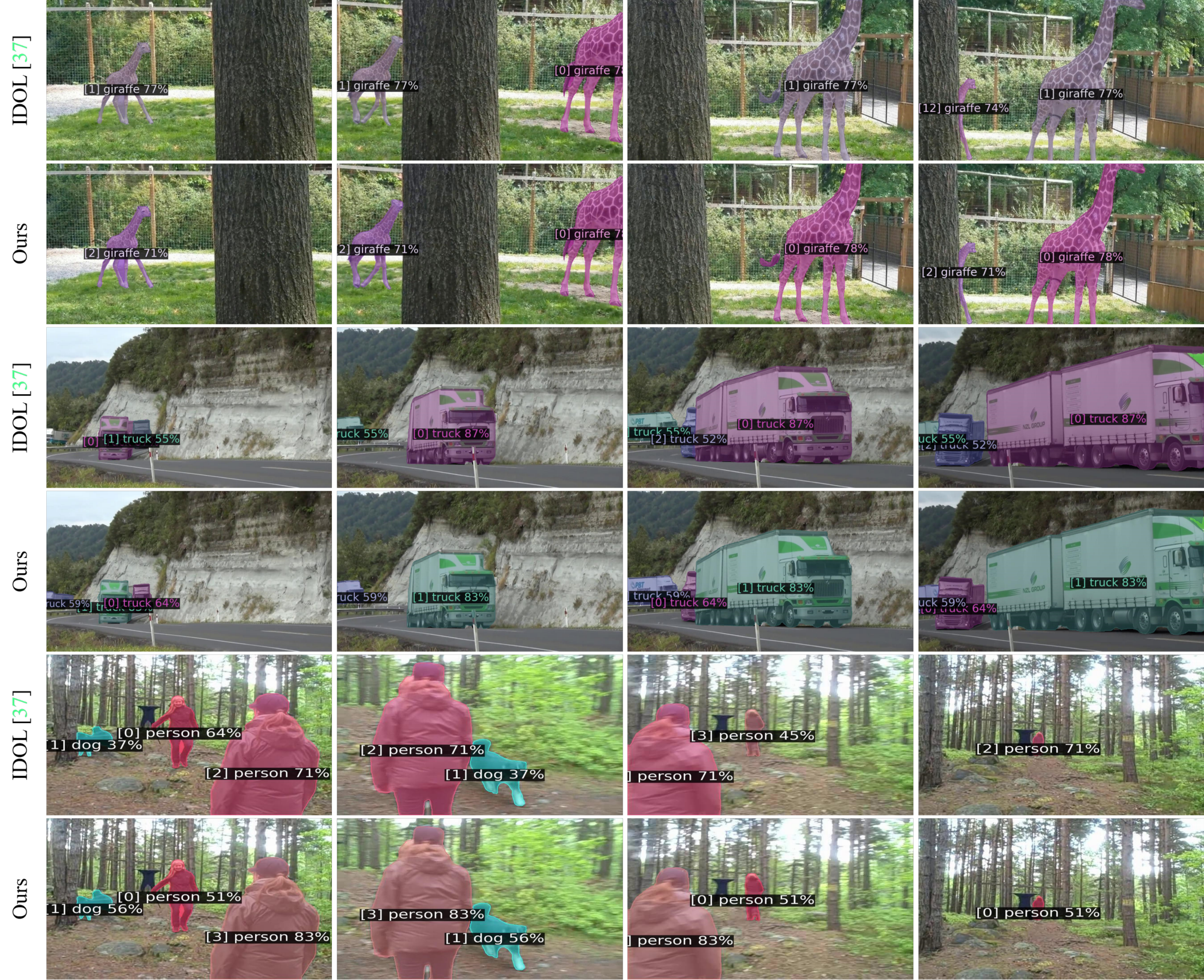}
  \caption{Qualitative comparison of our method to IDOL \cite{wu2022defense} for the YTVIS-22 dataset.}
  \label{fig:qualitative}
\end{figure*}

\textbf{YTVIS-19} is the first video instance segmentation dataset originated from Video Object Segmentation (VOS) datasets.
It includes 2,238 training, 302 validation, and 343 test data of high-resolution YouTube video clips.
They have 40 different object categories, and the video frame interval is 5.
YTVIS-19 has a small number of instances and a small amount of class variety for a single video (an average of 1.3 classes and 1.7 instances per video for the training set). 

\textbf{YTVIS-21} is an upgraded version of YTVIS-19, which has additional 747 training data and 119 validation data.
There are also 40 different object categories, but with some minor changes to the object class.
It includes a total of 2,985 training, 421 validation, and 453 test videos with an average of 1.5 classes and 3.4 instances per video for the training set.

\textbf{YTVIS-22} has the same training set as YTVIS-21, but 71 videos are additionally included in the YTVIS-21 validation set.
These additional videos, named `long videos', have longer frame intervals of 20, from longer video sequences (48.3 lengths for the long video and 31.3 lengths for the short video).
For clarification, the existing validation datasets are named `short video'; these have a frame interval of 5.

\textbf{OVIS} data is a very challenging dataset that contains long video sequences with a large number of objects and more frequent occlusion. 
It consists of 607 training, 140 validation, and 154 test videos. 
This dataset has a large number of instances despite its average class diversity (average l.4 class and 5.9 instances per video for the training set). This property makes VIS models more difficult to distinguish from each instance because each instance has a similar appearance.

\noindent\textbf{Evaluation Metric:} We use standard metrics for VIS, the average precision (AP), and average recall (AR) with the video intersection over Union (IoU) of the mask sequences as the threshold. 

\noindent\textbf{Baselines:} We use ResNet-50~\cite{he2016deep} and Swin-L~\cite{liu2021swin} backbones. ResNet-50 is the most standard and widely used backbone for VIS. Swin-L is a recent backbone that provides the best performance in VIS. 

\noindent\textbf{Implementation Details:} Unless otherwise noted, we follow the hyper-parameter setting of IDOL \cite{wu2022defense} for our online model and VITA \cite{heo2022vita} for offline knowledge aggregation.
We sample 4 frames to train the offline model in \secref{sec:offknowledge}.
We train the ResNet-50-based model in eight RTX3090 GPUs and the Swin-L-based model in eight A6000 GPUs. 
Our method using the Resnet-50 and Swin-L backbones runs at 30.6 fps and 17.6 fps for per-frame inference of the YTVIS-21 dataset, respectively.

\begin{table*}[h!]
\centering
\begin{tabular}{c | c | c c c c c c}
 \hline
 Type & Method         & AP   & AP$_{50}$ & AP$_{75}$ & AR$_{1}$ & AR$_{10}$ \\
\hline\hline
\multirow{3}{*}{Offline} & IFC \cite{hwang2021video}           & 13.1   & 27.8       & 11.6       & 9.4       & 23.9       \\
                         & SeqFormer \cite{wu2022seqformer}      & 15.1 & 31.9       & 13.8       & 10.4      & 27.1       \\
                         & VITA \cite{heo2022vita}          & 19.6 & 41.2       & 17.4       & 11.7      & 26.0        \\
\hline
\multirow{7}{*}{Online}  & M-RCNN \cite{yang2019video}        & 10.8 & 25.3       & 8.5        & 7.9       & 14.9       \\
                         & SipMask \cite{cao2020sipmask}        & 10.2 & 24.7       & 7.8        & 7.9       & 15.8       \\
                         & CrossVIS  \cite{yang2021crossover}     & 14.9 & 32.7       & 12.1       & 10.3      & 19.8       \\
                         & InstanceFormer \cite{koner2022instanceformer} & 20   & 40.7       & 18.1       & 12.0       & 27.1       \\
                         & DeVIS  \cite{caelles2022devis}        & 23.7 & 47.6       & 20.8       & 12.0       & 28.9       \\
                         & MinVIS \cite{huang2022minvis}      & 25.0              & 45.5         & 24.0 & 13.9    & 29.7 \\ 
                         & IDOL    \cite{wu2022defense}       & 30.2 & 51.3       & 30.0         & \textbf{15.0}       & 37.5       \\
                         & OOKD (Ours)          & \textbf{31.1} & \textbf{52.8}       & \textbf{32.7}       & \textbf{15.0}       & \textbf{39.6}        \\
 \hline
\end{tabular}
\caption{Quantative comparison of our method to state-of-the-art methods on the OVIS validation set. All results are conducted with ResNet-50 backbone. The best results are highlighted with \textbf{bold}.}
\label{table_OVIS}
\end{table*}

\begin{table}[]
\begin{tabular}{c|c|ccc}
\hline
{type} & Method      & {AP} & mAP$_S$ & mAP$_L$ \\
\hline\hline
\multirow{2}{*}{Offline} & Mask2former \cite{cheng2021mask2former} & 36.3 & 40.2 & 32.3 \\
                         & VITA  \cite{heo2022vita}                & 38.8 & 45.7 & 31.9    \\
                         \hline
\multirow{3}{*}{Online}  & MinVIS \cite{huang2022minvis}           & 34.5 & 43.5 & 25.6                       \\
                         & IDOL \cite{wu2022defense}               & 39.3 & 44.7 & 33.9                       \\
                         & OOKD (Ours)                     & \textbf{43.6} & \textbf{46.1} & \textbf{41.2}                       \\
\hline
\end{tabular}
\caption{Quantative comparison of our method to state-of-the-art methods on the YTVIS-22 dataset. All experiments are conducted with ResNet-50 backbone.}
\label{table_YTVIS_2022}
\end{table}

\subsection{Comparison to State-of-the-art Methods}


\textbf{YTVIS-21:} We conduct the performance comparison of our model to the recent competitive methods for the YTVIS-21 dataset in~\tabref{table_YTVIS_2021}.
Our method achieves the highest mAP for both ResNet-50 and Swin-L backbone by reaching mAP performances of 46.1\% and 59.2\%, respectively.
Compared to the state-of-the-art online model IDOL~\cite{wu2022defense}, the proposed method shows approximately 2\% and 3\% performance improvement on both backbones respectively.
Interestingly, OOKD outperforms the state-of-the-art offline model, VITA~\cite{heo2022vita}.

\textbf{YTVIS-22:} 
We compare our method to the most recent competitive methods~\cite{cheng2021mask2former,heo2022vita,huang2022minvis,wu2022defense} with the YTVIS-22 dataset.
We report quantitative results with mAP for short video (mAP$_S$) and long video (mAP$_L$) in \tabref{table_YTVIS_2022}.
Because the results for long videos are not reported in the papers, we use the source codes provided by the authors to measure the average precision in this experiment.
We achieve the best performance among all competitive methods in both short and long video datasets. 
We observe that the performance improvement for long videos is significant as ours outperforms the state-of-the-art offline (VITA) method with 9.3\% and the online (IDOL) method with 7.3\%. 
These results show that object-centric features associated with global video information extracted by OOKD enable robust feature matching between images with long time intervals.
Accurate feature matching is the key to increasing mAP scores. 

We also evaluate the methods qualitatively in \figref{fig:qualitative}.
All the competitive methods produce high-quality segmentation results while predicting wrong instance labels. 
This problem is significant, especially in the existing online-based methods.
It is because the inconsistent instance features per frame are extracted, and it makes the matching difficult. 
On the other hand, the proposed method correctly predicts the instance IDs although it is processed in an online manner.
The matching performance is improved by the proposed distillation method, which is the key to accurate video instance segmentation.

\textbf{OVIS:} 
The quantitative comparisons for the OVIS dataset are shown in \tabref{table_OVIS}.
The results also demonstrate that the proposed method outperforms all the competitive methods even with challenging datasets, OVIS.
It is widely known that offline VIS methods are struggling for video instance segmentation with long and dynamic videos \cite{heo2022vita,huang2022minvis}.
As is known, the experiments show that the recent online methods~\cite{wu2022defense,huang2022minvis,caelles2022devis} are generally better than the recent offline methods~\cite{heo2022vita,wu2022seqformer,hwang2021video}.
Despite the limitations of offline methods, our online method distilled by offline knowledge outperforms the state-of-the-art online method.
This demonstrates the effectiveness of offline knowledge distillation.
We believe that the instance features aggregated by global video information act as the proxies of each instance and the proxies guide all the features for the same instance to be consistent. 
This guideline helps the feature matching to be more robust even for the instances with appearance changes, and this is analyzed in detail in \secref{sec:ablation}.


\subsection{Ablation Study}
\label{sec:ablation}


\noindent\textbf{Knowledge Distillation with/without QFA}: To demonstrate the effectiveness of the query filtering and association (QFA), we conduct the ablation study in \tabref{tab:ablation_QFA}.
The results in the first and last rows are the results of the baseline model and our method, respectively.
The results in the second row are from a baseline model with the same knowledge distillation method without QFA-based query matching.
None of the results in \tabref{tab:ablation_QFA} adopts the data augmentation.
The results show that KD without QFA degrades the performance of the online model by about 0.4\% mAP$_{S}$ and 2.8\% mAP$_{L}$.
It is because the online model is trained by distilling mismatched offline knowledge into online features.
The order of instance IDs is not always consistent, and the instance labels should be matched to distill knowledge correctly.
Our model with KD and QFA improves the performance of the pure baseline model by approximately 1.0\% mAP$_{S}$ and 4.5\% mAP$_{L}$ thanks to the QFA finding corresponding features.
This shows that knowledge distillation is effective only with the proposed QFA. 
One interesting observation here is that the performance gains on long video datasets are significant.
This demonstrates that the offline-to-online knowledge distillation helps the online model to extract consistent features, even with the long video frames containing large appearance changes.


\noindent\textbf{Synergy of Minor-Paste and KD} 
We conduct the ablation study on Minor-Paste and KD in \tabref{tab:ablation_augKD}.
Minor-Paste improves the performance of the baseline model by 1.2\% and 0.9\% for the short video and the long video, respectively.
Moreover, our method with both augmentation and KD+QFA significantly improves the baseline model, especially for the long video datasets from 33.8\% to 41.2\%. 
Our KD+QFA brings 4.5\% out of a total of 7.4\% improvements and can be regarded as the core component that derives the success of our model.
Another interesting point here is the proposed augmentation scheme is more effective on our model than on the baseline model for the long video sequences (ours: 2.9\% and baseline: 0.9\% improvements). 
This shows that the richer knowledge of the minor class helps the knowledge distillation.
We also compare our Minor-Paste to the conventional copy-paste method~\cite{kim2022tubeformer} in \tabref{tab:ablation_aug}. 
As the proposed method shows better performance than the conventional method, mitigating the data imbalance problem helps better knowledge distillation.

\begin{table}[t!]
\centering
\begin{tabular}{cc|cc}
\hline
KD & QFA & mAP$_S$ & mAP$_L$ \\
\hline
X       & -   & 44.6    & 33.8    \\
O       & X   & 44.2    & 31.0    \\
O       & O   & 45.6    & 38.3   \\
\hline
\end{tabular}
\caption{Ablation study on knowledge distillation (KD) and KD with query filtering and association (QFA). We measure mAP$_{S}$ for short videos and mAP$_{L}$ for long videos in the YTVIS-22 dataset.}
\label{tab:ablation_QFA}
\label{table_QFA_ablation}
\end{table}

\begin{table}[t!]
\centering
\begin{tabular}{cc|cc}
\hline
Minor-Paste & KD+QFA & mAP$_S$ & mAP$_L$ \\
\hline
X   & X    & 44.6   & 33.8   \\
O   & X    & 45.8   & 34.7   \\
X   & O    & 45.6   & 38.3   \\
O   & O    & 46.1   & 41.2   \\
\hline
\end{tabular}
\caption{Ablation study on Minor-paste and our KD (KD+QFA).}
\label{tab:ablation_augKD}
\end{table}

\begin{figure}[t]
    \centering
    \includegraphics[height=0.64\linewidth]{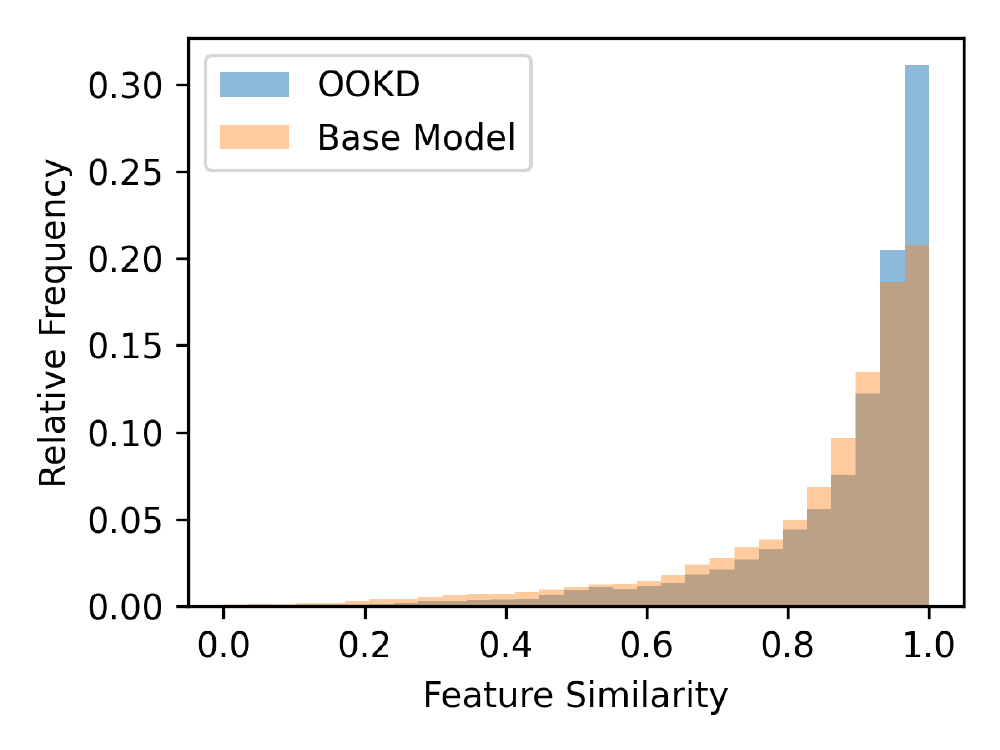}
    \caption{
    Histogram of feature similarity between two features from the same instance. The similarities of the features from our method and base model, IDOL~\cite{wu2022defense} are indicated by blue and orange bars, respectively.
    }
    \label{fig:pos_cos}
\end{figure}

\noindent\textbf{Instance Feature Similarity:} We perform further analysis of feature similarity to figure out why the proposed KD induces a performance improvement.
We report a histogram of the similarity between two features of the same instances that appeared at different frames in \figref{fig:pos_cos}. 
We randomly sample a pair of instances from all frames in 100 videos on the YTVIS-21 training dataset and measure cosine similarity.
The results show that higher feature similarity is obtained after OOKD is applied. 
The feature extraction with higher similarity among the same instances induces better performance. 
It is because online VIS models find instance IDs by selecting the instance with the highest feature similarity among all queries.
Thus, our model achieves better performance than the conventional method.

\begin{table}[t!]
\centering
\begin{tabular}{c|cc}
\hline
Method & mAP$_S$ & mAP$_L$ \\
\hline
KD without augmentation           & 45.6   & 38.3   \\
KD with Clip-Paste~\cite{kim2022tubeformer}        & 44.4   & 40.2   \\
KD with Minor-Paste (Ours)      & 46.1   & 41.2   \\
\hline
\end{tabular}
\caption{Ablation study on augmentation methods.}
\label{tab:ablation_aug}
\end{table}



\section{Conclusion}
\label{sec:conclusion}

In this paper, we propose OOKD, offline-to-online knowledge distillation for video instance segmentation.
Our method transfers the richer instance representation from an offline model into an online model.
To teach the student model correctly, we present a query filtering and association (QFA) that filters out irrelevant queries and finds the correct matching pairs between student and teacher queries.
This enables a single online model to take both advantages of online and offline models, which boosts the robustness while maintaining the ability for on-the-fly inference.
Robustness is further enhanced by our method of minor-paste augmentation that alleviates the class imbalance issues.
Extensive experiments have shown that our method improves the performance of the VIS, even in long and dynamic videos.
We also achieve state-of-the-art performance on YTVIS-21, YTVIS-22, and OVIS datasets by reaching mAP up to 46.1\%, 43.6\% and 31.1\%, respectively.

\clearpage
{\small
\bibliographystyle{ieee_fullname}
\bibliography{egbib}
}

\end{document}